\documentclass[10pt, a4paper]{article}

\usepackage[]{lrec-coling2024} 

\usepackage{xstring}

\usepackage{color}
\usepackage{array}
\usepackage{caption}
\usepackage{subcaption}
\usepackage{amsmath}
\usepackage{amsfonts}
\usepackage{amssymb}
\usepackage{bbm}
\newcommand\scalemath[2]{\scalebox{#1}{\mbox{\ensuremath{\displaystyle #2}}}}

\title{Mind Your Neighbours: Leveraging Analogous Instances for Rhetorical Role Labeling for Legal Documents}

\name{Santosh T.Y.S.S, Hassan Sarwat, Ahmed Abdou, Matthias Grabmair} 

\address{School of Computation, Information, and Technology; \\ Technical University of Munich, Germany
\\
\{santosh.tokala, hassan.sarwat,  ahmed.abdou, matthias.grabmair\}@tum.de\\}

\abstract{
Rhetorical Role Labeling (RRL) of legal judgments is essential for various tasks, such as case summarization, semantic search and argument mining. However, it presents challenges such as inferring sentence roles from context, interrelated roles, limited annotated data, and label imbalance. This study introduces novel techniques to enhance RRL performance by leveraging knowledge from semantically similar instances (neighbours). We explore inference-based and training-based approaches, achieving remarkable improvements in challenging macro-F1 scores. For inference-based methods, we explore interpolation techniques that bolster label predictions without re-training. While in training-based methods, we integrate prototypical learning with our novel discourse-aware contrastive method that work directly on embedding spaces. Additionally, we assess the cross-domain applicability of our methods, demonstrating their effectiveness in transferring knowledge across diverse legal domains.
 \\ \newline  }

\begin{document}

\maketitleabstract

\section{Introduction}

In an era of rapid digitalization and exponential growth of legal case volumes, the demand for automated systems to assist legal professionals in tasks like extracting key case elements, summarizing cases, and retrieving relevant cases has surged \cite{zhong2020does}. At the core of these tasks lies Rhetorical Role Labeling (RRL), which involves 
assigning functional roles to the sentences in the document such as preamble, factual content, evidence, reasoning, etc. Legal documents, characterized by their extensive length, lengthy sentences with unusual word order, frequent cross-references, extensive citation usage, and intricate lexicon, often feature uncommon expressions from everyday language and borrowed terms from various languages to the extent that they are referred to as a sub-language of legalese \cite{chalkidis2022lexglue,haigh2013legal}.


The task of RRL faces several distinctive challenges. Firstly, contextual dependencies, influenced by surrounding sentences and the case's context, are pivotal in discerning rhetorical role of each sentence, distinguishing RRL as a \textit{sequential sentence classification} task.
Secondly, the intertwining nature of rhetorical roles further complicates the task. For instance, the rationale behind a judgment (Ratio of the decision) often overlaps with Precedents and Statutes, necessitating a nuanced understanding of these roles' intricate distinctions \cite{bhattacharya2021deeprhole}.
Thirdly, obtaining extensive annotated data for specialized domains like law is expensive, requiring expert annotators. 
Lastly, certain rhetorical roles are disproportionately represented in the dataset, leading to significant class imbalance \cite{malik2022semantic,bhattacharya2021deeprhole}. Traditional up/down sampling methods struggle to fully address this challenge due to the task's nature, which involves sequences of sentences at the document level.

Initially RRL task is formulated as sentence classification, treating each sentence in isolation \cite{ahmad2020understanding,walker2019automatic}. Researchers later adopted it as sequential sentence classification, addressing contextual dependencies between sentences \cite{bhattacharya2021deeprhole,ghosh2019identification,malik2022semantic,kalamkar2022corpus}.
They introduced a two-level hierarchical model, encoding sentences independently at the lower level and contextualizing them with neighbouring sentences at the higher-level. While this approach effectively addressed the first challenge of RRL, other challenges remain unaddressed. 
Recently, \citealt{santosh2023joint} 
aimed to address data scarcity through data augmentation, but methods like word deletion, sentence swapping and back-translation could introduce noise and disrupt coherence. However, this approach did not effectively address label imbalance and intricate role intertwining.

In this work, we hypothesize that harnessing knowledge from semantically and contextually similar instances can provide valuable insights to grasp a broader context and reveal underlying rare patterns. This can enhance the understanding of complex label-semantics relationships, improve nuanced label assignments and equip the model to handle less common labels, thus addressing the distinctive challenges of RRL. We explore two approaches for harnessing this knowledge: one directly at inference time without additional parameters or re-training (Sec. \ref{RQ1}), and the other during training by incorporating auxiliary loss constraints (Sec. \ref{RQ2}). In the inference-based approaches, we interpolate the label distribution predicted by a model with the distribution derived from analogous instances in the training dataset, employing nearest neighbor-based, single, and multiple prototype-based methodologies. These methods enhance performance, particularly on more challenging macro-F1 scores, without requiring re-training. For training-based approaches, we integrate contrastive and prototypical learning which operate directly on the embedding space, leveraging neighborhood relationships. Additionally, we introduce a novel discourse-aware contrastive loss to address the contextual nature of the task. Our experimental results on four datasets from the Indian Jurisdiction validate our proposed methods.



While it is common to develop models for specific courts or domains due to unique vocabulary, complex linguistic structures and specific writing styles, such specialization can hinder the adaptability of these models beyond their original context. In rhetorical role labeling, models might memorize context-specific vocabulary rather than understanding the underlying semantics, making cross-domain applications challenging \cite{savelka2021lex}.
In such cases, developing a model for a new context typically requires annotating a new dataset, which can be expensive. In our work, we assess the cross-domain generalizability of our methods and observe that they enhance model's ability to transfer across different legal domains compared to a baseline model lacking these auxiliary techniques (Sec. \ref{RQ4}).

\section{Related Work}
\textbf{Rhetorical Role Labeling} Initial efforts of RRL aimed to facilitate summarization tasks \cite{saravanan2008automatic,farzindar2004letsum}. \citealt{saravanan2008automatic} employed Conditional Random Fields on hand-crafted features, to identify seven rhetorical roles in Indian state High Court documents. 
\citealt{savelka2018segmenting} categorized text into functional segments (Introduction, Background, Analysis, and Footnotes) and issue-specific segments (Analysis and Conclusion) using CRF on a corpus of US trade secret and cybercrime decisions.  \citealt{walker2019automatic} adopted feature-based methods for segmenting U.S. Board of Veterans’ Appeals decisions. \citealt{nejadgholi2017semi} focused on identifying factual and non-factual sentences in Canadian immigration case documents, using FastText embeddings for query-oriented search engine application. 

Recently, deep learning-based classification have been applied to this task in various contexts, such as Japanese documents \cite{yamada2019neural}, Indian Supreme Court documents \cite{bhattacharya2021deeprhole,ghosh2019identification,malik2022semantic,kalamkar2022corpus}. These methods adopt hierarchical approaches to account for the sequential sentence classification nature of the task, drawing context from surrounding sentences. This has been the defacto architecture for this task, with modifications ranging from word embeddings initially \cite{bhattacharya2021deeprhole,ghosh2019identification} to BERT based contextualized embeddings recently \cite{malik2022semantic,kalamkar2022corpus}. Recently, \citealt{santosh2023joint} reformulated the task as span-level sequential classification that segment the document into sets of contiguous sequence of sentences (spans) and assign them labels. In our work, we make use of the Indian Supreme Court corpus from prior research, proposing algorithms to effectively enhance their performance leveraging the knowledge from analogous neighbourly instances both at inference time without re-training, and also during training. 
Recently, \citealt{savelka2021lex} investigated the transferability of rhetorical segmentation models across seven jurisdictions and six languages, including Canada, the US, Czech Republic, Italy, Germany, Poland and France. In this study we also examine cross-domain performance on Indian Supreme Court documents across different legal contexts. 
\newline

\noindent \textbf{Leveraging Neighborhood Information} 
Utilizing neighborhood information offers two pathways: one during inference and the other during training. \textit{Inference-time methods}, commonly applied in few-shot classification, facilitate label assignment based on proximity to training examples without any re-training 
\cite{snell2017prototypical,yang2020simple}. Various techniques include employing all examples to identify nearest neighbors for the final label assignment \cite{yang2020simple} and constructing prototypes based on examples of the same label \cite{snell2017prototypical}, among others. This concept has gained widespread attention as retrieval-augmented models in various tasks, including language modeling \cite{khandelwal2019generalization,zheng2021adaptive}, machine translation \cite{zheng2021adaptive}named entity recognition \cite{wang2022k} and multi-label text classification \cite{wang2022contrastive}.

On the training side, methods like contrastive learning have been applied in self-supervised representation learning \cite{gao2021simcse}, wherein neighbour constraints are enforced in the embedding space through data augmentation. More recently, they have been extended to supervised learning scenarios using instances with the same label as neighbors \cite{khosla2020supervised}. Another approach is prototypical learning \cite{ding2020prototypical}, which designates representative prototypes for each class as guiding points to enforce neighborhood constraints on data instances. In this study, we harness both training and inference-based neighbour learning strategies. Additionally, we explore the 
their capabilities in cross-domain scenarios, within the context of the RRL task.

\section{Task, Datasets, Baseline}
\noindent \textbf{Task} Given a judgment document $x = \{x_1,x_2,\ldots,x_m\}$ with m sentences as the input, where $x_i = \{x_{i1},x_{i2},\ldots,x_{in}\}$ represents the $i^\text{th}$ sentence containing $n$ tokens and $x_{jp}$ refers to the $p^\text{th}$ token in the $j^\text{th}$ sentence, the task of rhetorical role labeling is to predict sequence of $l = \{l_1,l_2,\ldots,l_m\}$ where  $l_i$ is the rhetorical role corresponding to sentence $x_i$ and $l_i \in$ L  which is set of predefined rhetorical role labels. 
\newline

\noindent \textbf{Data} We experiment on four datasets - \textbf{(i) Build} \cite{kalamkar2022corpus} comprises judgments from Indian supreme court, high court, and district courts. It includes publicly available train and validation splits, with 184 and 30 documents respectively with a total of 31865 sentences (an average of 115
per document). These documents pertain to tax and criminal law cases and are annotated with 13 rhetorical role labels, including `None'. Given the absence of a public test dataset, we utilize the training dataset for both training and validation, evaluating performance on the validation partition. \textbf{(ii) Paheli} \cite{bhattacharya2021deeprhole} features 50 judgments from the Supreme Court of India across five domains: Criminal, Land and Property, Constitutional, Labour and Industrial, and Intellectual Property Rights, annotated with 7 rhetorical roles. They have total of 9380 sentences with an average of 188 per document. \textbf{(iii) M-CL / (iv) M-IT} \cite{malik2022semantic} encompasses judgments from the Supreme Court of India, High Courts, and Tribunal courts. It includes two subsets: M-CL, comprising 50 documents related to Competition Law, and M-IT, with 50 documents related to Income Tax cases. Both subsets are annotated with 7 rhetorical role labels. M-CL has 13,328 sentences (avg. of 266 per document) and M-IT has a total of 7856 sentences (avg. of 157 per document). We split (at document level) Paheli/M-CL/M-IT into 80\% train, 10\% validation, and 10\% test set. 
\newline

\noindent \textbf{Baseline}  All of our experiments in this study are built on top of the Hierarchical Sequential Labeling Network, which served as a baseline in prior works \cite{kalamkar2022corpus,santosh2023joint}. Initially, each sentence $x_i$ is encoded independently using a BERT model \cite{kenton2019bert} to derive token-level representations $z_i = \{z_{i1},z_{i2},\ldots,z_{in}\}$. These representations are passed through a Bi-LSTM layer \cite{hochreiter1997long}, followed by an attention pooling layer \cite{yang2016hierarchical}, to yield sentence representations $s = \{s_1, s_2,\ldots, s_m\}$. 
\begin{equation}
    u_{it} = \tanh(W_w z_{it} + b_w ) 
\end{equation}
\begin{equation}
    \alpha_{it} = \frac{\exp(u_{it}u_w)}{\sum_s \exp(u_{is}u_w)}  ~~\& ~~ 
    s_i = \sum_{t=1}^n \alpha_{it}u_{it}
\label{att1}
\end{equation}
Here, $W_w$, $b_w$, and $u_w$ are trainable parameters. The sentence representations $s$ are passed through Bi-LSTM layer to obtain contextualized representations $c = \{c_1, c_2, \ldots, c_m\}$ that encode contextual information from surrounding sentences. Finally, the contextual representations $c$ 
are passed through a Conditional Random Field layer that predicts the best sequence of labels. 


\section{RQ 1: Leveraging the Neighbourhood at Inference}
\label{RQ1}

In this section, we leverage the knowledge from semantically similar training instances directly during inference without extra training overhead. We interpolate the label distribution predicted by the baseline model with the distribution derived from the training instances similar to the test instance. 
This overcomes the problem of memorizing/learning rare patterns implicitly in the model parameters, thus enhancing the model's ability to handle long-tail cases (classes with few instances or rare patterns in frequent classes) especially in limited data settings. 
We explore three different methods to obtain the distribution from similar training instances.

\subsection{Methods}
\subsubsection{Interpolation with kNN}
In this method, we construct a datastore of training instances and then retrieve the nearest neighbours to the test instance for computing the interpolated label distribution during the inference. 
\newline 

\noindent \textbf{Datastore Construction} After training, we obtain contextualized representation $c_i$ of every sentence in each document of the training set using the trained model. We construct the datastore by a single forward pass over each training document. The datastore $\{K, V\}$ is the set of all contextualized representation-rhetorical label pairs constructed from all the training examples $D$ as:
\begin{align}
    \{K,V\} = \{(c_i,l_i)| \forall x_i \in x, \forall l_i \in l, (x,l) \in D\}
\end{align}

\noindent \textbf{Interpolation} 
During inference time, we query the datastore using the contextualized representation of every sentence in the test document,  to find the k-nearest neighbours $N$ according to the euclidean distance. Then, we derive the distribution of labels $p_{kNN}$ using labels of the retrieved neighbours based on softmax of their negative distances, while aggregating probability mass for each label across all its occurrences in the retrieved neighbours (labels that do not appear in the retrieved $N$ have zero probability). Intuitively, the closer a neighbor is to the test instance, the larger its weight is. 
\begin{equation}
    p_{kNN}(l_i|x, x_i) \propto \sum_{(k,v)\in N} \mathbbm{1}_{l_i = v}
\exp( \frac{-d(c_i, k)}{\tau})
\end{equation}
$\tau$ denotes the temperature hyperparameter and d(.) indicates euclidean distance. Finally, we interpolate the $p_{baseline}(l_i |x, x_i )$ with $p_{kNN}(l_i |x, x_i)$ as:
\begin{align}
\begin{split}
p_{final}(l_i|x, x_i) = {}& \lambda p_{baseline}(l_i|x, x_i) + \\ & 
(1 -\lambda) p_{kNN}(l_i|x, x_i)
\end{split}
\label{interpolation}
\end{align}

\noindent where $\lambda$ makes a balance between derived $p_{kNN}$ and  $p_{baseline}$ obtained from the trained model.

\subsubsection{Interpolation with Single Prototype} 

Instead of storing all the training instances in the datastore, we seek to store one prototype for each label, which captures the essential semantics of various sentences for each rhetorical role, significantly reducing the datastore's memory footprint. To create a prototype for each label, we calculate the average of the contextualized embeddings of sentences that share the same rhetorical role. 
Intuitively these prototypes can be assumed to be the center of clusters for different labels, surrounded by sentences expressing the same label in the embedding space. The interpolation process closely resembles the kNN approach (Eq. \ref{interpolation}), with the key difference being that interpolation directly involves the prototypes, rather than 
a prior retrieval step. 

\subsubsection{Interpolation with Multiple Prototypes}

Instead of using a single prototype for each rhetorical role, we suggest the use of multiple prototypes for each label. This choice is driven by the fact that instances with the same rhetorical role can exhibit distinct variations in expression, resulting in diverse contextual embeddings scattered across the embedding space. Averaging these embeddings into a single prototype might diminish specificity. Utilizing multiple prototypes allows us to effectively capture the intricate viewpoints within each label.
To accomplish this, we cluster the instances belonging to each rhetorical role using $k$-means, yielding multiple prototypes for each label from the k centroids. The interpolation step remains similar (Eq. \ref{interpolation}), involving all these multiple prototypes without any retrieval step.

\subsection{Experiments}
\subsubsection{Implementation Details} \label{impl-rq1} We follow the hyperparameters for baseline as described in \citealt{kalamkar2022corpus}. We use the BERT base model to obtain the token encodings. We employ a dropout of 0.5, maximum sequence length of 128, LSTM dimension of 768, attention context dimension of 200. We sweep over learning rates \{1e-5, 3e-5, 5e-5. 1e-4, 3e-4\} for 40 epochs with Adam optimizer \cite{kingma2014adam} to derive the best model based on validation set performance. For all our inference variants, we carry a grid search over the interpolation factor ($\lambda$) in increments of 0.1 in the range of [0,1] to choose the best model based on Macro-F1 on validation set. For KNN and multiple prototypes, we vary k over powers of 2 from 8 till 256. 

\begin{table*}[]
\begin{tabular}{|l|c|c|c|c|c|c|c|c|}
\hline
\textbf{}               & \multicolumn{2}{c|}{\textbf{Build}}                  & \multicolumn{2}{c|}{\textbf{Paheli}}                 & \multicolumn{2}{c|}{\textbf{M-CL}}                   & \multicolumn{2}{c|}{\textbf{M-IT}}                   \\ \hline
\textbf{}               & {\textbf{mac.F1}} & \textbf{mic.F1} & {\textbf{mac.F1}} & \textbf{mic.F1} & {\textbf{mac.F1}} & \textbf{mic.F1} & {\textbf{mac.F1}} & \textbf{mic.F1} \\ \hline
\textbf{Baseline}       & {60.20}           & 79.13          & {62.43}          & 66.02          & {59.51}          & 67.04          & {70.76}          & 70.50          \\ \hline
\textbf{+ KNN}          & {62.92}          & 81.04          & 66.53 & 70.82  & {63.14}          & 73.02          & {72.16}          & 71.62          \\
\textbf{+ Single Proto} & {61.23}          & 80.12          & {62.43}          & 66.02          & {61.42}          & 71.64          & {71.97}          & 71.08          \\ 
\textbf{+ Mutli Proto}  & {63.23}          & 81.96          & {65.36}          & 70.02          & {62.73}          & 72.78          & {72.82}          & 72.46          \\ \hline
\end{tabular}
\caption{Performance of interpolation methods on four datasets. mac.F1: macro-F1, mic.F1: micro-F1}
\label{tab-RQ1}
\end{table*}

\subsubsection{Results} 
In Table \ref{tab-RQ1}, we present the macro-F1 and micro-F1 scores for both the baseline and the interpolation variants. We observe a significant improvement when using kNN interpolation across all datasets, particularly in the more challenging macro-F1 metric, which accounts for label imbalances. 
On the other hand, single prototype interpolation mitigates memory footprint issue of kNN by storing one representation per rhetorical role but leads to performance degradation compared to kNN.
This decline results from oversimplification, as a single prototype may struggle to capture the diverse aspects within each rhetorical role, particularly when instances of the same label are dispersed across the embedding space. 
This is evident in the Paheli dataset, where no improvement over the baseline is observed. 
Interpolation with multiple prototypes balances memory efficiency and label variation capture.
While it slightly underperforms kNN interpolation in Paheli and M-CL datasets, it outperforms kNN in Build and M-IT. This can be 
attributed to a smoothing effect that reduces noise or human label variations in the kNN-based approach, particularly evident in datasets with low inter-annotator agreements (Build and M-IT). 
These results affirm our hypothesis that straightforward interpolation using training set examples during inference can boost the performance of rhetorical role classifiers.
\newline 

\begin{figure}
\centering
    \includegraphics[width=0.9\linewidth]{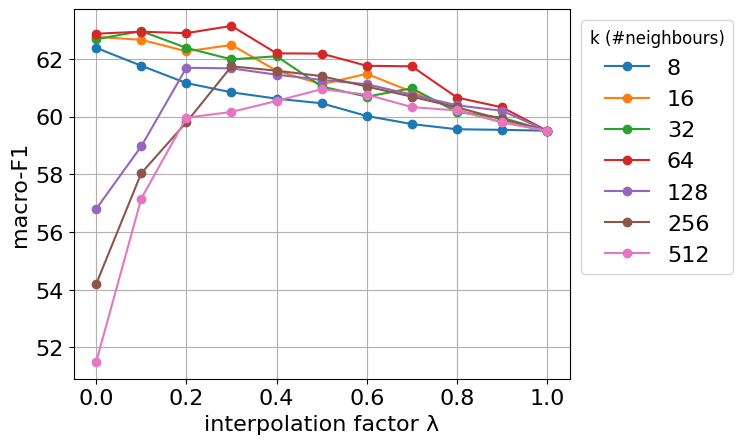}
    \caption{Sensitity to hyperparameters - kNN (M-CL) $\lambda$ = 0: interpolation only, $\lambda$ = 1: baseline only}
    \label{knn-hyperparam}
\end{figure}

\noindent \textbf{Sensitivity of interpolation} In Figure \ref{knn-hyperparam}, we present the macro-F1 score for the M-CL dataset using kNN interpolation, while varying the interpolation coefficient $\lambda$ and the number of neighbors 'k'. Here, $\lambda$ values of 0 and 1 correspond to predictions solely from interpolation and the baseline model, respectively. 
We observe that performance initially improves as 'k' increases, signifying that incorporating more neighbors boosts confidence by including closely similar examples. However, performance starts to decline with higher 'k',
which can be attributed to a large number of neighbours introducing noise with low inter-annotator agreement, suggesting a need for a addressing this task a multi-label classification.
On the other hand, reducing $\lambda$ consistently enhances performance, particularly for lower k, showcasing the model's capacity to rely solely on semantically similar instances for label prediction. With higher k, we notice a decline in performance at lower $\lambda$ values beyond a certain optimal point, which is related to the label variation problem exacerbated by a larger number of neighbours. Similar trends are observed with other interpolations. 

\section{RQ 2: Leveraging the Neighbourhood at Training}
\label{RQ2}


We leverage the knowledge from neighbour instances directly at the training process to improve the performance. We explore three methods: contrastive learning, single prototypical learning, and multi prototypical learning. 
These techniques draw inspiration from the same principles as their inference-time counterparts but serve as auxiliary loss constraints during training. Their primary aim is to improve the discriminative capability of embeddings by highlighting differences between instances with distinct rhetorical roles and similarities among instances sharing the same label.

While the task-specific classification loss focuses on mapping contextualized representations to label outputs with supervision on individual instances, the methods in this section directly operate on embeddings in latent space. They exploit the interplay among instances to establish effective discriminative decision boundaries, serving as a form of regularization.

\subsection{Methods}
\subsubsection{Contrastive Learning}
Contrastive learning aims to bring an anchor point closer to related samples while pushing it away from unrelated samples in the embedding space. In a supervised setting, samples with the same/different labels are considered related/unrelated with respect to an anchor \cite{khosla2020supervised}. The loss is calculated as follows:
\begin{align}
     L^{cont =- \frac{1}{N^2} \sum_{i,j}} & \frac{\exp(\delta(c_i, c_j )d(c_i, c_j ))}{\sum_{j'} \exp(1 - \delta(c_i, c_{j'}))d(c_i, c_{j'} )}  \\
     d(c_i,c_j)  & =\frac{1}{(1+ \exp( \frac{c_i}{|c_i|} · \frac{c_j}{|c_j|}))}
\label{contrastive}  
\end{align}
\noindent where $\delta(c_i, c_j )$ denotes 1 if $c_i$ and $c_j$ have same rhetorical label, else 0, N denotes batch size.


Lengthy legal documents limits the number of documents that can be accommodated in a single batch and this raises concerns about having enough positive samples for the minority class instances within a batch for effective contrasting. To overcome this limitation, we utilize a \textbf{memory bank} \cite{wu2018unsupervised}, where we progressively reuse encoded representations from previous batches to compute the contrastive loss. In practice, we maintain a fixed-size representation queue for each rhetorical role. As new representations corresponding to specific labels are generated, they are enqueued into the respective queue with their gradients detached. If the queue size for a label exceeds the maximum limit, the oldest element is dequeued. When computing the contrastive loss, we use the same equation \ref{contrastive}. However, in addition to the current batch instances, we employ all the representations stored in the memory bank for contrasting purposes, using them as both positives and negatives, based on the anchor point's label.

\begin{table*}[]
\scalebox{0.95}{
\begin{tabular}{|l|c|c|c|c|c|c|c|c|}
\hline
\textbf{}                                                                          & \multicolumn{2}{c|}{\textbf{Build}}                    & \multicolumn{2}{c|}{\textbf{Paheli}}                   & \multicolumn{2}{c|}{\textbf{M-CL}}                     & \multicolumn{2}{c|}{\textbf{M-IT}}                     \\ \hline
\textbf{}                                                                          & {\textbf{mac.F1}} & \textbf{mic.F1} & {\textbf{mac.F1}} & \textbf{mic.F1} & {\textbf{mac.F1}} & \textbf{mic.F1} & {\textbf{mac.F1}} & \textbf{mic.F1} \\ \hline
\textbf{Baseline}                                                                  & {60.20}            & 79.13           & {62.43}           & 66.02           & {59.51}           & 67.04           & {70.76}           & 70.50            \\ \hline
\textbf{+ Contrastive}                                                             & {64.55}           & 83.54           & {68.06}           & 71.91           & {62.24}           & 72.42           & {73.41}           & 73.53           \\ 
\textbf{+ Contrastive + MB}                                                        & {66.51}           & 83.29           & {71.76}           & 72.69           & {63.14}           & 72.72           & {72.22}           & 72.46           \\ 
\textbf{+ Disc. Contr.}                                                            & {66.37}           & 83.81           & {71.99}           & 73.85           & {66.94}           & 73.02           & {72.23}           & 74.01           \\ 
\textbf{+ Disc. Contr. + MB}                                                       & {66.48}           & 83.67           & {71.19}           & 73.28           & {64.72}           & 72.36           & {72.85}           & 73.05           \\ \hline
\textbf{+ Single Proto.}                                                           & {66.01}           & 81.45           & {69.94}           & 71.09           & {64.42}           & 71.52           & {72.59}           & 71.98           \\ 
\textbf{+ Multi Proto.}                                                            & {66.35}           & 83.05           & {71.38}           & 72.92           & {65.91}           & 73.57           & {73.02}           & 74.13           \\ \hline
\textbf{\begin{tabular}[c]{@{}l@{}}+ Disc. Contr. \\ + Single Proto.\end{tabular}} & {67.02}           & 83.91           & {74.28}           & 73.86           & {65.87}           & 72.12           & {72.50}            & 72.1            \\ 
\textbf{\begin{tabular}[c]{@{}l@{}}+ Disc. Contr. \\ + Multi Proto.\end{tabular}}  & {67.21}           & 83.65           & {75.52}           & 76.34           & {68.66}           & 74.59           & {73.14}           & 72.22           \\ \hline
\end{tabular}}
\caption{Results on four datasets for methods leveraging neighbourhood during training (RQ2). Contr., Disc., MB, Proto. indicates Contrastive, Discouse-aware, Memory Bank and Prototypical respectively.  }
\label{tab-RQ2}
\end{table*}


To incorporate the concept of context from surrounding sentences into contrastive learning, we introduce a novel \textbf{discourse-aware contrastive loss}. This is based on the idea that sentences in close proximity within a document, sharing the same label, should exhibit a stronger proximity in the embedding space compared to sentences with the same label but positioned farther apart in the document. To implement this concept, we introduce a penalty inversely proportional to the absolute difference in their positions. In particular, we impose a higher penalty on positive sentence pairs that are closer in the document, encouraging them to be closer in the embedding space than pairs originating from greater distances within the document. The discouse-aware loss is as follows:


\begin{align}
\scalemath{0.95}{L^{cont}  = - \frac{1}{N^2} \sum_{i,j}} & \scalemath{0.95}{\frac{\exp( \beta(i,j) \delta(c_i, c_j )d(c_i, c_j ))}{\sum_{j'} \exp(1 - \beta(i,j) \delta(c_i, c_{j'}))d(c_i, c_{j'} )}}  \\
\beta(i,j)  \propto  & \frac{1}{|j-i|}
\end{align}
\noindent where $\beta$ represents a penalty that considers positional information. When $c_i$ and $c_j$ come from different documents, such as cross-document positives/negatives from the memory bank or across the batch, we apply the lowest possible penalty, considering $c_i$ as the farthest sentence relative to in-document positives. We incorporate this additional contrastive loss alongside the classification loss during training. 

\subsubsection{Single Prototypical Learning}
While contrastive learning guides instances to adjust their positions in the embedding space relative to other instances 
, prototypical learning employs specific prototypes for each label in the embedding space which act as specific guiding points \cite{ding2020prototypical}.
Specifically, we randomly initialize one prototype for each label as $z = \{z_1, z_2, \ldots, z_k\}$, where $k$ denotes the number of rhetorical roles. 
These prototypes undergo fine-tuning during the training and we apply distance-related constraints from both the prototype's and the sample's perspectives to guide their relationships.
(i) Prototype centric view (pcv): aims to bring samples $S_j$ belonging to label $j$ closer to the corresponding prototype $z_j$, while simultaneously pushing away samples of other labels, denoted as $S_j'$, from this prototype. (ii) Sample centric view (scv): In this view, the representation $c_j$ with label $j$ is drawn closer to its designated prototype $z_j$, while pushing away from other prototypes $Z_j' = z - z_j$. These two views are represented in loss as:
\begin{equation}
    \scalemath{0.8}{
    L^{pcv}_j  = - \frac{1}{N} (\sum_{c_p \in S_j}  \log (d(z_j,c_p)) + \sum_{c_i \in S_j'} \log(1-d(z_j,c_i)))}
\end{equation}
\begin{equation}
\scalemath{0.85}{
    L^{scv}_j  = - \frac{1}{K} (  \log (d(z_j,c_j))+ \sum_{z_p  \in Z_j'} \log(1-d(z_p,c_j)))}
\end{equation}
\noindent 
These both views shape the embedding space by aligning prototypes with their corresponding samples, forming distinct clusters of different labels, each centered around a specific prototype vector.

\subsubsection{Multi Prototypical Learning}
Instead of using a single prototype for each label, this approach employs multiple prototypes for each label to capture the diverse variations within the sentences of the same label. To implement this, a set of $M$ prototypes  per label is randomly initialized and a diversity loss \cite{zhang2022protgnn} is integrated to penalize prototypes of the same label if they are too similar to each other.
This ensures that prototypes of the same label are distributed across the embedding space, capturing the multifaceted nuances 
under each label. The Sample Centric View is also modified to ensure that each sample is in close proximity to at least one prototype among all the prototypes of the same class.
\begin{align}
\begin{aligned}
L^{div}_k & =  \sum_{\substack{q\neq r \\ z_q,z_r \in Z_k}} \max(0, z_q \cdot z_r - \theta) 
\end{aligned} \\
\begin{aligned}
L^{scv}_j &  = -  \min_{z_q \in Z_k }  \log (d(z_q ,c_j) + \\
 & \frac{1}{(k-1)M}{\sum_{z_p  \in Z_k'}} \log(1-d(z_p,c_j))
\end{aligned}
\end{align}
\noindent where $z_q$, $z_r$ are prototypes of same label $k$. 
Sample $c_j$ belongs to label k. ${\theta}$ is the similarity threshold. 
$Z_k$ and $Z_k'$ represent the set of prototypes corresponding to the label $k$ and those of labels other than k, respectively.

\begin{figure*}
    \centering
    \begin{subfigure}{0.22\textwidth}
        \includegraphics[width=\linewidth]{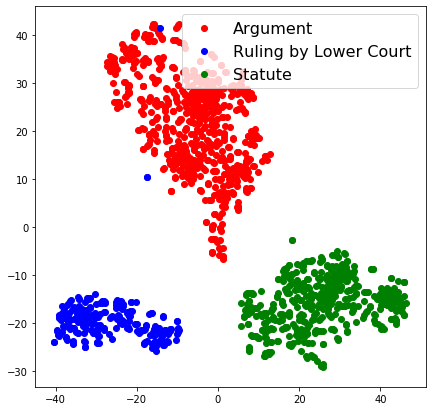}
        \caption{Contrastive}
        \label{cont-vis}
    \end{subfigure}
    \hfill 
    \begin{subfigure}{0.22\textwidth}
        \includegraphics[width=\linewidth]{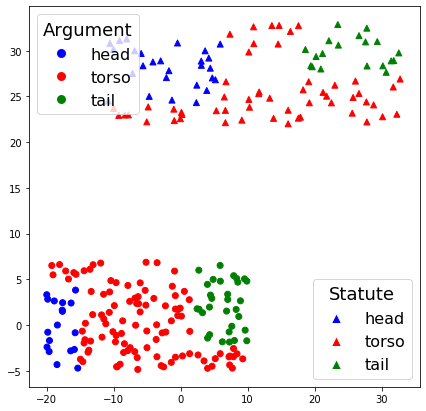}
        \caption{Disc.-aware Contr.}
        \label{weig-cont-vis}
    \end{subfigure}
    \hfill
    \begin{subfigure}{0.22\textwidth}
        \includegraphics[width=\linewidth]{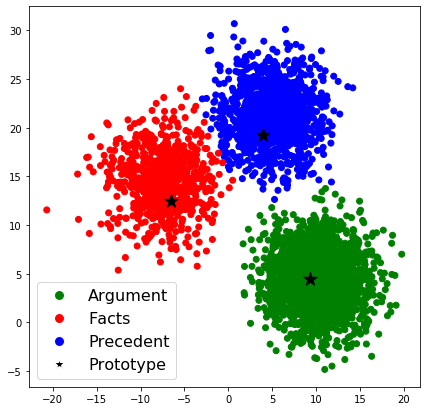}
        \caption{Single-Prototypical}
        \label{single-proto-vis}
    \end{subfigure}
    \hfill
    \begin{subfigure}{0.22\textwidth}
        \includegraphics[width=\linewidth]{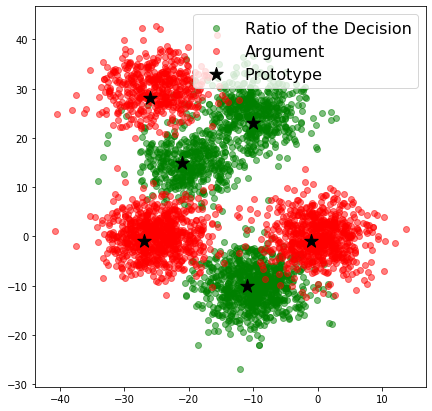}
        \caption{Multi-prototypical}
        \label{multi-proto-vis}
    \end{subfigure}
    \caption{t-SNE visualizations of different models on M-CL dataset. Disc.: Discourse, Contr.: Contrastive. \\ head, torso and tail in Disc.-aware Contr. plot indicate the relative position of the sentence in a document.}
    \label{visual}
\end{figure*}

\subsection{Experiments}
\subsubsection{Implementation Details}
We use the same training setup as described in Sec. \ref{impl-rq1}. We conduct grid-search for size of memory bank per label and number of prototypes in multi-prototypical learning in powers of 2 from [32,512] and [4,256] respectively using the validation set performance. 

\subsubsection{Results} 
Table \ref{tab-RQ2}, shows that incorporating contrastive loss improves performance across all datasets. Furthermore, the discourse-aware contrastive loss, which leverages relative position to organize embeddings, enhances performance, supporting our hypothesis that sentences with the same label and in close proximity in the document should be closer in the embedding space. 
Augmenting the contrastive loss with a memory bank further enhances performance, particularly in macro-F1, benefiting sparse classes. However, the degree of improvement is less or negative in the discourse-aware variant. This can be due to the positional factor, as additional sentences from other documents retrieved from the memory bank are placed at the end of the document, leading to smaller penalization factors and contributing only marginally to the loss. Overall, the discourse-aware contrastive model emerges as the most effective among the contrastive variants.

The single prototypical variant performs comparably to the best contrastive variant and outperforms the baseline. This demonstrates that specific guiding points through prototypes can effectively aggregate knowledge from neighboring instances. Moreover, multiple prototypes further improve performance, highlighting the need to capture multifaceted nuances. These results suggest that the addition of respective losses can eliminate the need to design specific memory banks to expose the model to large batches for effective guidance from neighbors in contrastive learning.


Finally, combining the discourse-aware contrastive variant with both single and multiple prototype variants yields further improvement, highlighting the complementarity between these approaches. These results suggest that deriving supervisory signals from interactions among training instances can be an effective strategy for addressing the class imbalance problem, particularly in low-data settings. 
\newline


\noindent \textbf{Qualitative Analysis: } To examine the impact of our auxiliary loss functions on the learned representations, we employ t-SNE \cite{hinton2002stochastic} to project the high-dimensional latent space hidden states obtained by the model in Fig. \ref{visual}. In the case of contrastive learning, we observe that sentences with the same label form distinct clusters. With the addition of discourse-aware contrastive loss, samples with the same label in a specific document adhere to the positional constraint, aligning with our hypothesis that samples sharing a label and closer in the discourse sequence should be positioned closer in the embedding space compared to those farther apart. In single prototypical learning, prototypes occupy the centers of corresponding sentences, forming distinctive manifolds. Similarly, multi-prototypical learning captures multifaceted aspects with prototypes dispersed across the embedding space, each prototype serving as the center for respective samples. These visualizations affirm the effectiveness of our learning methods. 


\section{RQ 3: Cross-Domain Generalizability}
\label{RQ4}
To evaluate how well our proposed methods can transfer across different domains, we train the model on one dataset (source) and assess its performance on other datasets (target) in a blind zero-shot manner. We use the Paheli, M-CL, and M-IT datasets, which span diverse domains but share a same 7 rhetorical label space. The resulting Macro-F1 scores are presented in Table \ref{tab-RQ4}.

\begin{table}[]
\scalebox{0.9}{
\begin{tabular}{|l|l|c|c|c|}
\hline
\textbf{Train \textdownarrow}          & \multicolumn{1}{|r|}{\textbf{Test \textrightarrow}}      & \textbf{Paheli} & \textbf{M-CL} & \textbf{M-IT} \\ \hline
 &  \textbf{Random}      &  19.10 & 7.87
& 9.12 \\ \hline
\textbf{Paheli} & Baseline       & \color[HTML]{9B9B9B}62.43           & 56.98         & 57.31         \\ 
\textbf{}       & Disc. Contr.   & \color[HTML]{9B9B9B}71.99           & 56.54         & 57.40          \\ 
\textbf{}       & Single Proto.  & \color[HTML]{9B9B9B}69.94           & 58.30          & 59.92         \\ 
\textbf{}       & Multi Proto.   & \color[HTML]{9B9B9B}71.38           & 57.47         & 59.48         \\ 
\textbf{}       & DC + Single Pr & \color[HTML]{9B9B9B}74.28           & 62.27         & 60.33         \\ 
\textbf{}       & DC + Multi Pr  & \color[HTML]{9B9B9B}75.52           & 60.89         & 60.61         \\ \hline
\textbf{M-CL}   & Baseline       & 54.71           & \color[HTML]{9B9B9B}59.51         & 63.08         \\ 
\textbf{}       & Disc. Contr.   & 54.04           & \color[HTML]{9B9B9B}66.94         & 62.98         \\ 
\textbf{}       & Single Proto.  & 57.48           & \color[HTML]{9B9B9B}64.42         & 60.23         \\ 
\textbf{}       & Multi Proto.   & 56.10            & \color[HTML]{9B9B9B}65.91         & 61.62         \\ 
\textbf{}       & DC + Single Pr & 59.95           & \color[HTML]{9B9B9B}65.87         & 63.92         \\ 
\textbf{}       & DC + Multi Pr  & 57.89           & \color[HTML]{9B9B9B}68.66         & 62.37         \\ \hline
\textbf{M-IT}   & Baseline       & 52.97           & 58.83         & \color[HTML]{9B9B9B}70.76         \\ 
\textbf{}       & Disc. Contr.   & 51.89           & 57.16         & \color[HTML]{9B9B9B}72.23         \\ 
\textbf{}       & Single Proto.  & 51.57           & 58.58         & \color[HTML]{9B9B9B}72.59         \\ 
\textbf{}       & Multi Proto.   & 52.85           & 58.70          & \color[HTML]{9B9B9B}73.02         \\ 
\textbf{}       & DC + Single Pr & 51.03           & 57.23         & \color[HTML]{9B9B9B}72.50          \\ 
\textbf{}       & DC + Multi Pr  & 51.77           & 56.99         & \color[HTML]{9B9B9B}73.14         \\  \hline
\end{tabular}}
\caption{Macro-F1 scores of our methods across three datasets. The column `train' indicates the source dataset on which the model is trained and each of the dataset columns indicates the target test dataset. Scores in grey indicates the in-domain performance (trained and tested on same dataset). \{DC, Disc. Contr.\} : Discourse-aware contrastive, \{Pr., Proto.\} : Prototypical }
\label{tab-RQ4}
\end{table}

Naturally, models trained and tested on the same domain outperform those trained on different domains
(e.g., baseline model trained and tested on Paheli achieves a Macro-F1 of 62.43, whereas trained on M-CL and tested on Paheli achieves 54.71). Interestingly, the baseline model shows an ability to transfer knowledge  from one domain to another, outperforming random\footnote{Random choices are based on the training set's label distribution (uniform distribution lead to further lower scores). These are averaged over 10 runs.} guessing across all datasets. While discourse-aware contrastive model improves in-domain performance, it marginally reduces cross-domain performance across all datasets compared to the baseline (e.g., Disc. Contr. trained on M-CL and tested on Paheli achieves a Macro-F1 of 54.04, while the baseline with the same setup achieves 54.71). This can be attributed to the model capturing domain-specific features while minimizing distances between similar instances in contrastive learning. In contrast, single and multi-prototypical models enhance cross-domain transfer compared to the baseline, except when trained on M-IT. This indicates prototypical learning acts as a more robust guiding point, preventing overfitting to noisy neighbors as in contrastive models. Between the two, single prototype tend to perform better, due to its single representation being agnostic to domain-specific variations and encapsulating core characteristics, making it more adept in cross-domain scenarios. Furthermore, coupling discourse-aware contrastive with prototypical models boosts cross-domain performance, except when trained on M-IT. This behaviour of M-IT may be attributed to marginal in-domain improvements, leading to overfitting on domain-specific features limiting cross-domain generalization. This prompts questions about selection of optimal source dataset for improved performance on target datasets, warranting further investigation. For instance, to test on Paheli with baseline, training on M-CL yields a Macro-F1 of 54.71, while on M-IT yields 52.97. Additionally, exploring joint training with multiple datasets could shed light on their impact on both in-domain source and unseen target datasets.

\section{Conclusion}

In this paper, we have demonstrated the potential for enhancing the performance of rhetorical role classifiers by leveraging knowledge from neighbours, semantically similar instances. Interpolation with kNN and multiple prototypes at the inference time have shown promising improvements, especially in addressing the challenging issue of label imbalance, without requiring re-training. Additionally, our approach of incorporating neighbourhood constraints during training with our proposed discourse-aware contrastive learning and prototypical learning has demonstrated improvements. Combining both methods has boosted it further, indicating their complementary nature. Notably, the prototypical methods have proven to be robust, showcasing performance gains even in cross-domain scenarios, generalizing beyond the domains they were trained on.

\section{Limitations}
One constraint in the current task formulation is that it restricts assigning a single label to each sentence, which may not fully account for the complexity of lengthy sentences that can encompass multiple rhetorical roles. To address this limitation, an alternative approach could involve reformulating the task as multi-label classification, enabling each sentence to be associated with more than one gold-standard rhetorical role. Another avenue for exploration is to shift from sentence-level segmentation towards a finer-grained approach at the phrase or sub-sentence level, necessitating the assignment of rhetorical roles to each phrase or sub-sentence while specifying the dependency relations between these segments \cite{tokala2023label}. 

It's important to acknowledge that while our cross-domain experiments have provided valuable insights into model generalizability, these evaluations have primarily focused on datasets originating from Indian courts, covering various domains within this single jurisdiction. The observed improved performance across these datasets could potentially be attributed to shared country-specific vocabulary and legal conventions. To ensure the robustness and generalizability in a broader context, it is imperative to expand the assessment to encompass diverse legal contexts across different countries and regions, 
where legal documents from  may exhibit significant linguistic and structural variations.

\section{Ethics Statement}
The scope of this work is to provide new methods along with corresponding experiments to drive research forward in rhetorical role labeling, which is a pivotal task constituting the inaugural step in the legal document processing pipeline. Our experiments have been carried out on four publicly available datasets from different Indian courts. Though these decisions are not anonymized and contain the real names of the involved parties, we do not foresee any harm incurred by our experiments. We believe that our research contributes positively to the broader goals of advancing legal NLP and the development of AI-driven tools for legal professionals. By enhancing the automation of rhetorical role labeling, we can streamline legal document analysis and significantly benefit the legal field.

\section{Bibliographical References}\label{sec:reference}

\bibliographystyle{lrec-coling2024-natbib}
\bibliography{lrec-coling2024}


\end{document}